\documentclass[letterpaper, times, psfig, 10 pt, conference]{ieeeconf}
\IEEEoverridecommandlockouts
\usepackage[backend=biber,
            hyperref=true,
            url=false,
            isbn=false,
            doi=false,
            backref=false,
            style=ieee,
            natbib=true,
            citestyle=numeric-comp,
            sorting=none,
            block=none]{biblatex}
            
\renewcommand{\bibfont}{\small}
\usepackage{multicol}
\usepackage{multirow}

\usepackage{graphics}
\usepackage[pdftex]{graphicx}

\usepackage[font={small}]{caption}
\usepackage{subcaption}

\usepackage{mathtools}
\usepackage{amsmath, amssymb, amscd}
\usepackage{wasysym} 
\usepackage{nicefrac}       
\usepackage{mdwmath}
\usepackage{mdwtab}
\usepackage{eqparbox} 
\usepackage{algorithm} 
\usepackage{algorithmicx, algpseudocode} 

\usepackage{array} 
\usepackage{tabularx}
\usepackage{multicol}
\usepackage{booktabs}

\usepackage[T1]{fontenc} 
\usepackage[utf8]{inputenc}
\usepackage[english]{babel} 
\usepackage{units}
\usepackage{bm}
\usepackage{xspace}
\usepackage{flushend}
\usepackage{balance}

\usepackage{csquotes}
\usepackage{makeidx}

\let\labelindent\relax
\usepackage{enumitem}

\usepackage{soul} 
\usepackage{subfiles} 

\usepackage[yyyymmdd]{datetime}

\date{\protect\formatdate{1}{1}{2001}}

\usepackage{url}
\usepackage{color}
\usepackage[usenames,dvipsnames]{xcolor}

\makeatletter
\g@addto@macro{\UrlBreaks}{\UrlOrds}
\makeatother

\usepackage{marginnote}
\usepackage{soul} 

\newcommand{\tocite}[1]{%
\textcolor{red}{[cite:\ifthenelse{\equal{#1}{}}{}{#1}?]}
}

\newcommand{\ignore}[1]{}

\algnewcommand{\IIf}[1]{\State\algorithmicif\ #1\ \algorithmicthen}
\algnewcommand{\EndIIf}{\unskip\ \algorithmicend\ \algorithmicif}

\usepackage{lipsum}

\usepackage{color,soul}

\newcolumntype{L}[1]{>{\raggedright\let\newline\\\arraybackslash\hspace{0pt}}m{#1}}
\newcolumntype{C}[1]{>{\centering\let\newline\\\arraybackslash\hspace{0pt}}m{#1}}
\newcolumntype{R}[1]{>{\raggedleft\let\newline\\\arraybackslash\hspace{0pt}}m{#1}}

\begin{document}

\title{\Large \bf  Semantic and Geometric Modeling with Neural Message Passing\\ in 3D Scene Graphs for Hierarchical Mechanical Search}

\author{
Andrey Kurenkov$^{1}$, 
Roberto Mart{\'i}n-Mart{\'i}n$^{1}$, %
Jeff Ichnowski$^{2}$,
Ken Goldberg$^{2}$,
Silvio Savarese$^{1}$
\thanks{
\scriptsize{$^{1}$Stanford University}
}
\thanks{
\scriptsize{$^{2}$University of California, Berkeley}
}
}

\maketitle

\begin{abstract}
Searching for objects in indoor organized environments such as homes or offices is part of our everyday activities. When looking for a desired object, we reason about the rooms and containers the object is likely to be in; the same type of container will have a different probability of containing the target depending on which room it is in. We also combine geometric and semantic information to infer what container is best to search, or what other objects are best to move, if the target object is hidden from view. We use a 3D scene graph representation to capture the hierarchical, semantic, and geometric aspects of this problem. 
To exploit this representation in a search process, we introduce \textit{Hierarchical Mechanical Search (HMS)}, a method that guides an agent's actions towards finding a target object specified with a natural language description. HMS is based on a novel neural network architecture that uses neural message passing of vectors with visual, geometric, and linguistic information to allow HMS to process data across layers of the graph while combining semantic and geometric cues.
HMS is trained on 1000 3D scene graphs and evaluated on a novel dataset of 500 3D scene graphs with dense placements of semantically related objects in storage locations, and is shown to be significantly better than several baselines at finding objects. It is also close to the oracle policy in terms of the median number of actions required. Additional qualitative results can be found at \url{https://ai.stanford.edu/mech-search/hms}
\end{abstract}


\IEEEpeerreviewmaketitle

\section{Introduction}
\label{sec:intro}
In our day-to-day lives we often face the task of looking for a particular object in organized indoor spaces. When searching for an object, the process is coarse-to-fine, sequential, and interactive: we first consider the room, then the container the object can be in, and then, after opening the container, we may have to move objects to be able to see the target. The combination of these steps makes this a hierarchical instance of mechanical search~\cite{danielczuk2019mechanical}. 

For example, when looking for cheese in a new house, we would look 1) in the kitchen, 2) within the fridge, and 3) near or behind other dairy products that may be occluding it, even if bigger but less related objects (e.g. a big watermelon) occlude a larger area. Further, while we expect the cheese to be in a fridge, we would not first search in a fridge in the garage, because that fridge is more likely to contain other items such as soda. In other words, each decision is informed by multiple levels (e.g. we expect different content for the same container in different rooms), and the search process is guided by both semantic and geometric reasoning of the target, rooms, containers, and occluding objects. 

\begin{figure}[!t]
\centering
\vspace{-5pt}
    \includegraphics[width=0.98\linewidth]{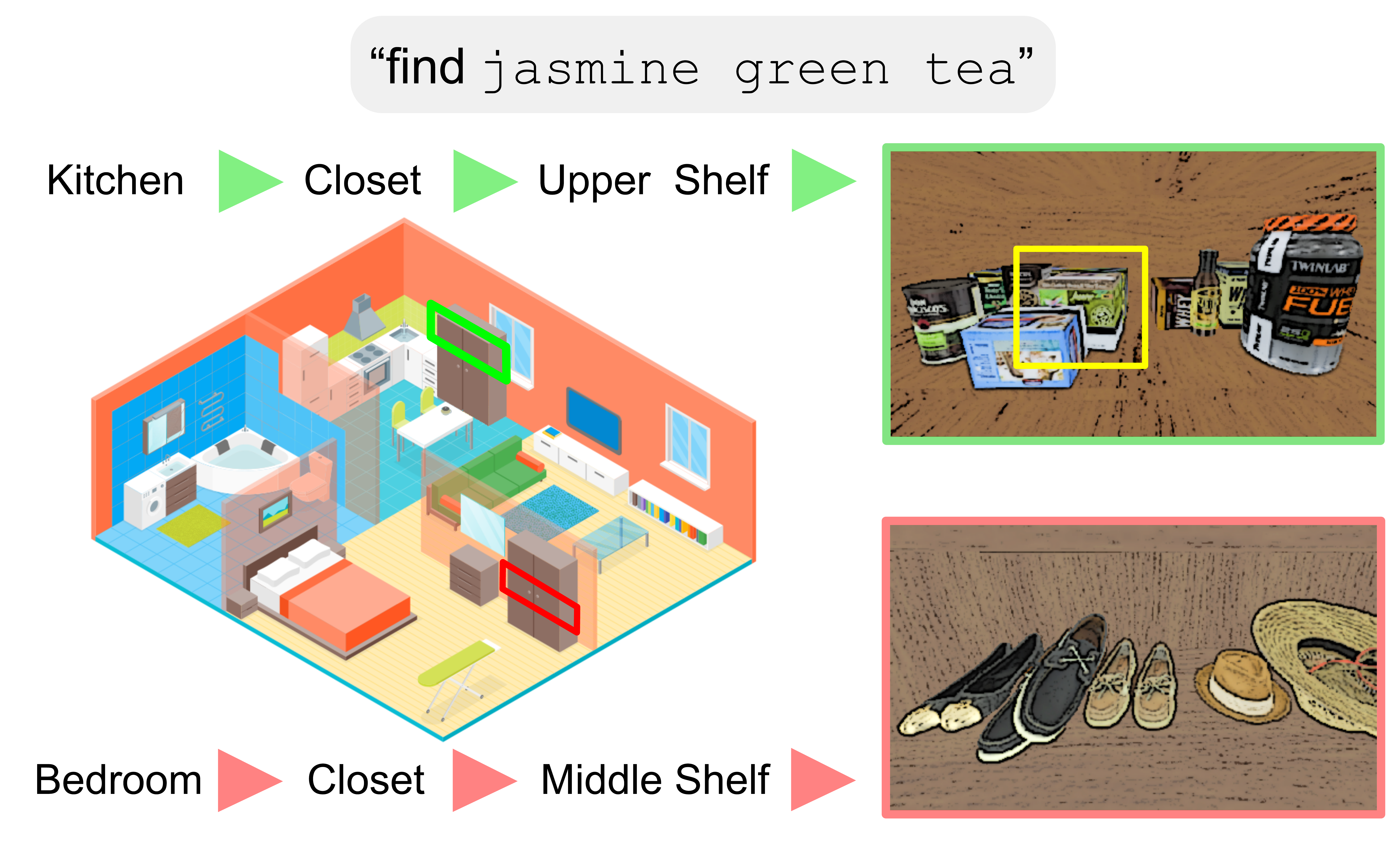}
    \caption{\textbf{Searching for an object in large environments using semantic and geometric information:} assuming knowledge of rooms and containers, the agent is tasked to find a target object (\texttt{jasmine green tea}). From coarse to fine location levels, the agent decides what room, container, and section of the container to search, and, if the target is not visible, what visible object should be removed to reveal it. \textit{Right top/bottom}: Content of the same container in different rooms. Information from different levels (e.g. room) helps predict and select at others (e.g. container); \textit{Right top image}: Target (yellow box) is partially behind semantically related objects that need to be removed before larger unrelated objects. The mechanical search at the shelf level is guided by a combination of geometric and semantic information.}
  \label{fig:intro-fig}
  \vspace{-15pt}
\end{figure}

In this work, we address the problem of searching for objects in hierarchical organized indoor environments (Fig.~\ref{fig:intro-fig}). The target object is specified by a short natural language description.
We assume that the searching agent is provided with the layout of the environment and the storage locations that can contain target objects, as well as visual observations of the objects inside a container the first time the agent explores it.
The goal of the agent is to find the target with a minimum number of actions (motion to a new room, selection of a new container, and/or removal of a possibly occluding object).
To efficiently search in this context, the agent needs to exploit information at different levels (room, container, object) while fusing geometric (size) and semantic (linguistic description and labels, appearance) information.

Previous learning-based approaches enabled agents to search for objects given images of the environment. However, these approaches reason only about semantics~\cite{yang2018visual, wu2019bayesian, qiu2020target, crespo2020semantic} or geometry~\cite{li2016act, xiao2019online, danielczuk2020x} of the objects at a single level, and ignore the combined nature of the search problem in organized environments. Furthermore, these works have assumed that the agent starts with little or no prior knowledge about the environment, which limits their ability to search as efficiently as possible when such prior knowledge is available. Several works in robotics~\cite{wong2013manipulation, moldovan2014occluded} have addressed joint reasoning over both semantic and geometric aspects of objects during search, but they are restricted to only one room or container, do not enable search for objects guided by text descriptions, and do not learn to reason about object occlusion from images; thus, they scale poorly to real world scenarios. 

In this work, we propose to represent the environment with an extension of 3D Scene Graphs~\cite{armeni20193d} that captures the hierarchical structure of the environment as well as the geometric and semantic information about containers and objects within it.
This graphical structure can represent rooms, containers and objects as a hierarchy of nodes that can grow dynamically as new objects are seen by the searching agent when it first searches inside object containers.

To enable an agent to search with joint reasoning about layers of the graph, as well as semantic and geometric information associated with nodes, we propose \textit{Hierarchical Mechanical Search (HMS)}. HMS uses neural message passing to propagate information in the form of feature vectors from individual nodes across the graph to be spread out to multiple other nodes. Feature vectors are created by converting labels into word vectors and appearance into the vectors output by a convolutional neural network, and then concatenating geometric values such as location and size to these vectors. Given a large dataset of examples, the model is trained to assess the likelihood of the target object to be in a given room node, inside a given container node, or behind a given object node. HMS also includes a search procedure that uses these likelihoods to take actions that will retrieve the target object. 
The neural message model is trained using a large number of procedurally generated 3D Scene Graphs, created from realistic probabilities of objects to be stored together in containers, and containers to be within rooms. This process also generates  realistic placements of object models inside containers to generate training images (Fig.~\ref{fig:intro-fig}, right).

In summary, the contributions of this work are:
\begin{itemize}[wide, labelwidth=!, labelindent=0pt]
    \item We introduce the problem of object search in organized indoor environments with partially known layout information and dynamically growing object information,
    \item We propose \textit{Hierarchical Mechanical Search (HMS)}, a search procedure guided by neural message passing that can be recursively applied on 3D scene graphs to jointly reason across layers of the graph and from semantic and geometric cues, which enables an agent to efficiently find target objects,
    \item We experimentally demonstrate that HMS significantly outperforms baselines and ablations on 500 unique household scene graphs unseen during training, which have 26 different object categories and 164 unique object types.
\end{itemize}

\section{Related Work}
\label{sec:rw}

Using real-world \textbf{semantic statistics of objects to guide search} has been explored in the past, for example through learning probable spatial relations between the target object and other objects in the environment~\cite{sjoo2012topological, aydemir2011search, toro2014probabilistic, lorbach2014prior, kunze2014bootstrapping, kim2019active,zeng2020semantic,crespo2020semantic}. 
More recent approaches applied Deep Reinforcement Learning to generate policies that search for a target using image inputs, assuming access to object detectors. These works have typically embedded co-occurence statistics between objects into graph structures that are converted into a vector using graph neural networks~\cite{yang2018visual, wu2019bayesian, qiu2020target,druon2020visual}.
Surveys of these works can be found in~\cite{crespo2020semantic} and \cite{zeng2020survey}.
Differently, we integrate hierarchical information of geometric and semantic cues, extracting the statistics of object co-occurence directly from training data. We also use the information differently: we actively guide an agent to select the rooms and containers to search in as well as the occluding objects to remove.


\begin{figure}[!t]
\centering
\vspace{-5pt}
    \includegraphics[width=0.98\linewidth]{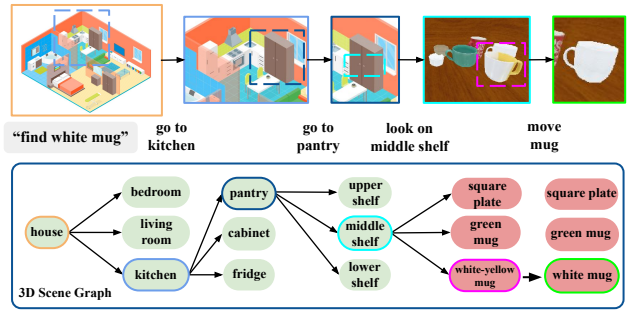}
    \caption{\textbf{Hierarchical semantic and geometric search:} for a given environment (top left) and a description of a target object (middle left), the agent is tasked with reasoning about an associated 3D scene graph (bottom) to reason about which containers (green circles) the object is likely to be in as well as which objects (red circles) it is likely to be occluded by, so as to move occluding objects out of the way and make the target object visible (right). 
    }
  \label{fig:overview-fig}
  \vspace{-15pt}
\end{figure}

There are other approaches to guide robot \textbf{manipulation with the environment} to try to make an occluded target object visible. Most of these works reason about geometry instead of semantics to infer the possible location of an object, and decide what objects should be moved using graph-based planning~\cite{dogar2014object, nam2019planning} or POMDP planning approaches~\cite{li2016act, xiao2019online}. We do not plan based on detailed geometric models, but instead leverage  coarser hierarchical geometric information together with semantic information to guide the search. Recent works used neural networks to visually reason about piles of objects and learn how to manipulate them with the objective of finding a target object~\cite{danielczuk2019mechanical,danielczuk2020x,kurenkov2020visuomotor,cheng2018reinforcement, novkovic2020object, boroushaki2020robotic}. Different to ours, those methods manipulate the pile and do not leverage semantic cues. Our work most closely relates to Wong et al.~\cite{wong2013manipulation}. They presented a POMDP approach that reasons jointly over object semantics and geometric properties to guide manipulation actions. However, in our approach we do not assume a static set of object types and instead learn to perceive how similar an object is to the target text description,  and also address the problem of jointly reasoning over multiple layers of information. 


To model our problem, we extend the \textbf{3D Scene Graph} representation~\cite{armeni20193d}  (Fig.~\ref{fig:overview-fig}). While there are many approaches for scene graph generation from images~\cite{xu2017scene,herzig2018mapping,yang2018graph}, our focus is on leveraging an already-generated scene graph to guide an agent in object search. Prior work has used similar graph architectures to reason about possible location of objects. Zhou et al.~\cite{zhou2019scenegraphnet} presented a \textbf{neural message passing} method to infer the most probable missing object in a 3D indoor scene based on the visible objects. While our approach also involves message passing on a scene graph, we apply this differently to reason jointly over multiple layers of the scene graph and about both semantic and geometric cues information, and to guide mechanical search.

\section{Problem Formulation}
\label{s_pf}

In this work, we assume the search task to be defined by a text description of the target object.
Text descriptions are unique characterizations of instances of a class, e.g. \texttt{garlic} is an instance of class \texttt{Veggies} and \texttt{Krill\_Oil\_Pills\_Blue} an instance of class \texttt{Vitamins}.
We assume the environment to be composed of two types of elements, containers and objects.
In this formulation, rooms, furniture, compartments inside furniture, \ldots are considered containers, which contain other containers or objects.
We also assume that the searching agent is provided with prior information about the containers: the house rooms and the furniture in each room (e.g. cupboards, drawers, fridges, \ldots), along with their semantic class and size. This is a realistic assumption, since the robot could just first visit each room to acquire such layout information. 
To make the problem more realistic, we assume that the content of the containers (i.e. objects) is not known until the agent decides to search them for the first time. Once a container is searched, the agent receives an image of its content and a simulated object detector will provide the \textbf{class label} (not text description) and the corresponding image bounding box around the objects that are at least 70\% unoccluded. Objects that are more than 70\% are not visible until the agent removes the objects occluding them.

When searching for a target object, the agent has two types of actions that it can take: 1) explore a container, or 2) remove an object to reveal what is behind it. Moving to a room, opening a cupboard, looking at a shelf or picking an object are counted as individual actions.
The goal of the agent is to take the minimum number of actions to find the target object in the environment.
In this work, we do not directly address the challenges of low-level robot control for navigation, grasping and manipulation of the objects; we assume the agent is able to move to desired locations and move desired visible objects once it chooses to do so. 

\begin{figure}[!t]
\centering
\vspace{-5pt}

    \begin{subfigure}{0.4\textwidth}
       \centering
    \includegraphics[width=\linewidth]{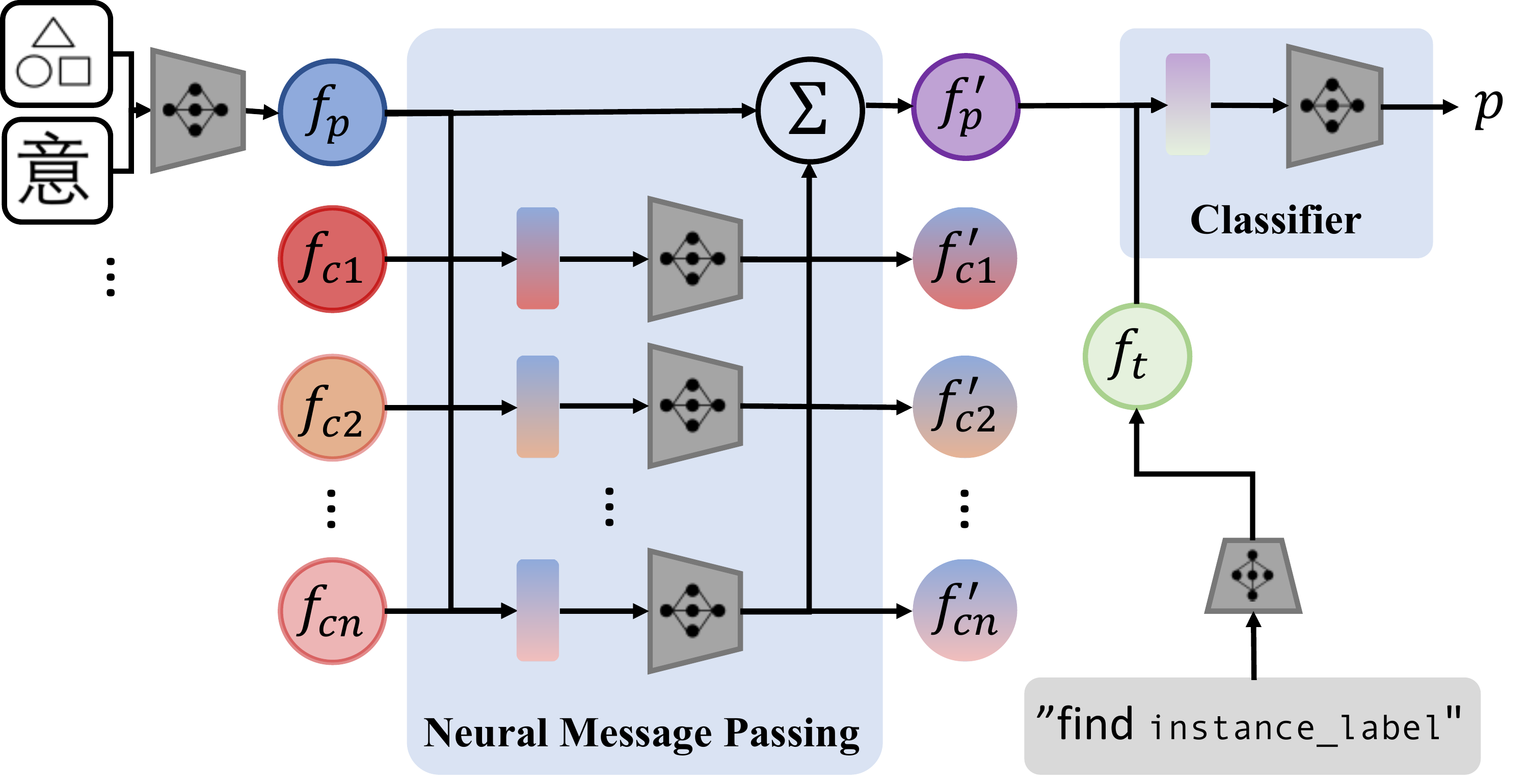}
    \end{subfigure}
    \caption{The neural net architecture we use recursively to search for a target object. Scene graph nodes are represented as circles, with their colors representing the contents of their feature vectors. Trapezoidal shapes correspond to neural net layers. Message passing is performed across neighboring layers in the graph, such that the parent node's feature vector is updated to reflect all of its children, and the children's nodes are updated to reflect their parent. As a result, the model is able to jointly reason across layers of the graph. }
  \label{fig:model-fig}
  \vspace{-15pt}
\end{figure}

\section{Hierarchical Mechanical Search}
\label{sec:approach}

We propose to represent the hierarchical structure of containers and objects with their semantic and geometric properties with a modified 3D Scene Graph~\cite{armeni20193d}. 
Our 3D Scene Graph (Fig.~\ref{fig:overview-fig}) represents the house's rooms and storage locations as nodes annotated with geometric and semantic information: the sizes of containers and labels indicating their semantic class (e.g. \texttt{kitchen}, \texttt{fridge}).
When a new container is explored and some objects in its interior are found and perceived by the provided detector (the ones that are at least 70\% visible), we extend the Scene Graph to include them and their information: their semantic class (e.g. \texttt{Veggies}, \texttt{Medicine}, \texttt{Hats}), their normalized size in the image (size of the detector's given bounding box), their normalized location in the image, and their appearance (an RGB image cropped to primarily show the object).

We now need an algorithmic structure that can reason on this hierarchical 3D Scene Graph and help the agent decide on the most optimal next action to take to find the target object. To address this, we introduce \textit{Hierarchical Mechanical Search (HMS)}, a search procedure that is guided by a novel neural message passing mechanism that enables integrating information across levels (see Fig.~\ref{fig:model-fig}). As part of this, we create feature vectors for each of the nodes in the graph (containers and objects) which can be used as input to our neural message passing architecture. These feature vectors should contain both the semantic and geometric information known about each node (Sec.~\ref{ss_fni}). Then, we can propagate this information across layers with the message passing model and infer the probability of containers to enclose the target, and of objects to occlude it (Sec.~\ref{ss_nmpm}). This model is optimized in an end-to-end supervised fashion with a classification objective (Sec~\ref{ss_opt}). After it is optimized, the agent's searching actions will follow the HMS search procedure based on these probabilities output by the model (Sec.~\ref{ss_sp}). 

\vspace{-2pt}
\subsection{Featurizing Node Information}
\label{ss_fni}

Nodes in the scene graph include geometric and semantic information of containers and objects. This information needs to be ``featurized'', or converted into a numeric vector, $f$, to be used as inputs to our model and propagated among levels of the graph to make informed decisions. Therefore, we first compute a feature vector with the information of each node, and of the target object, $f_t$.

Semantic information in the form of language labels (i.e.  \texttt{bedroom},  \texttt{fridge}) is featurized using the corresponding pre-trained 300-dimensional GloVe word vector~\cite{pennington2014glove}; word vectors are known to encode semantic meanings, and so aid in associating the target object's description with the appropriate containers and objects. The feature vector of the target object $f_t$ is created by averaging the GloVe vectors of all the words in its description. For containers, the Glove vector is concatenated with their known volume.

In the case of objects, the information that needs to be featurized includes their label, bounding box size, bounding box location, and the RGB image of the object. We featurize the label in the same way as done for containers, and concatenate it to the normalized center coordinates and size of the bounding box. To be able to reason about the appearance of the object, we use the bounding box coordinates to obtain an image of the object and crop it to show the object with some of the container surroundings around it in the container also visible. For extracting features from this image, we fine-tune the resnet-18 Convolutional Neural Net~\cite{he2016deep}  from weights pretrained on the ImageNet classification task~\cite{deng2009imagenet}, after replacing its final layers with two randomly initialized fully connected layers that output a 256-dimensional vector that can be trained for our task.

Lastly, all vectors are processed by two fully connected layers to obtain final feature vectors of the same dimension of 100 for containers, the target, and non-target objects.

\vspace{-2pt}
\subsection{Neural Message Passing Model}
\label{ss_nmpm}

Given the scene graph and the feature vectors of nodes and the target object, we propose a single neural net architecture that can be trained to evaluate each container for whether it contains the target object, and each detected object for whether it is occluding the target object. To evaluate a single container in a way that makes use of information about subsequent layers in the graph (e.g. its contained objects), we utilize \textit{neural message passing}, a technique for neural-net based reasoning on graph structures that propagate information between nodes. Message passing is a better choice for our problem compared to other forms of learning on graph structures because it can be applied on dynamically changing graph structures, and on graphs of different sizes and depths. 

Our message passing mechanism is depicted in Fig.~\ref{fig:model-fig} and summarized as follows: first, we combine the parent's (a container) feature vector, $f_p$, with the feature vectors of each of its $n$ children nodes (other containers or objects), $f_{c1},\ldots,f_{cn}$, by concatenating them, and further process the combinations with learnable fully connected layers. This results in a new set of children vectors, $f'_{c1},\ldots,f'_{cn}$, that contain information about both the parent and the child. These new vectors with parent information are stored as the child's new feature vector to use in later processing. All of the new child vectors are aggregated via averaging, and a new parent feature vector, $f'_{p}$, is made by averaging the original vector with the aggregated parent-child vector. The updated parent feature vector combines the information of the parent and its children (e.g. both a room and all the containers within it). 


The new parent feature vector, or the stored feature vector of a leaf object node, is combined with the target object's vector, $f_t$, by element-wise multiplication. This combined representation encodes information about both the candidate node and the target. The combined vector is passed through three fully connected layers, followed by a sigmoid activation. The final output, $p$, a value between 0 and 1, is optimized to predict the likelihood of the target object to be in the considered container, or behind the considered object. While different classification layers are trained for evaluating containers and objects, the architecture is exactly the same for both, resulting in a recursive repeating pattern at each parent-child level of the 3D Scene Graph hierarchy.

\vspace{-2pt}
\subsection{Model Optimization}
\label{ss_opt}
In order to utilize the model for search, it needs to be trained to output correct probabilities, $p$. We will train our neural message passing architecture using a large dataset of scene graphs of indoor environments with annotated locations for target objects in a supervised manner. The procedural generation of this dataset using annotations of plausible object locations is  described in Sec.~\ref{sec:scenegraphs}. At each optimization step, the following process is followed: 

\begin{enumerate}[wide, labelwidth=!, labelindent=0pt]
    \item We sample batches of scene graphs from the train dataset. 
    \item For each scene graph, we sample an occluded object and set it to be the target object for that scene graph.
    \item For each layer in each scene graph, we sample one node having a positive label (i.e. a container that has the target object, or an object that occludes the target object) and one node having a negative label with respect to the target object.
    \item We compute and store feature vectors for all sampled nodes and their children.
    \item The model computes probabilities $p$ for all of sampled nodes, starting with the top-most layer and proceeding to the bottom-most layer. Computation has to proceed in this way, since for a given parent node $n_p$, the updated child representations $f'_{c1},\ldots,f'_{cn}$ are used as input to the model when computing outputs in the children's layer.
    \item For each sampled node, we use its known positive or negative label and the model's output $p$ to compute the binary cross entropy classification loss.
    \item  We average all losses and optimize all model parameters (resnet-18, ``featurization'', message passing, and classification layers) using the Adam optimizer~\cite{kingma2014adam} end-to-end.
\end{enumerate}



\subsection{Search Procedure}
\label{ss_sp}

The above neural net architecture is used to guide a procedure similar to greedy depth-first-search for finding the target object. This is chosen over a variant of breadth-first-search to mimic the manner in which an agent would explore an environment, with the agent not being able to 'teleport' and therefore exploring all or most of a given container before proceeding to the next most promising option.

The search proceeds in the following way: starting at the top-most layer of the graph, we separately evaluate each node for whether it is likely to contain the target object. We then sort the nodes by the model's output, and search each one in turn by evaluating how likely its children nodes are to contain or occlude the target. Each child node is evaluated using the same message passing procedure as in the first layer, with the childrens' starting feature vectors, $f'_{c1},\ldots,f'_{cn}$, carried over from the message passing step done with the parent. Once evaluated, the children nodes are again searched in sorted order, until they are all explored and the search reverts to the next node in the layer above. For containers, ``search'' means deciding which of the containers or objects within that container to search, whereas for objects ``search'' means moving the object out of the scene.

An issue of the above approach is that it would search nodes even if the model gives them a low probability $p$ of containing or occluding the target object. So as to avoid this, for all nodes if $p$ is lower than a threshold $T$, the node is not searched but rather skipped. The value of $T$ starts the same for all nodes, but can be updated during the search so as to prevent the correct nodes for finding the target object from never being searched. This update happens whenever a node is skipped, at which point the value of $T$ for all subsequent nodes in that layer is set to to the value of $p$ of that node. If the target is not found the first time all top-level nodes are either searched or skipped, the procedure restarts with the updated values of $T$ and removed objects from the previous iteration being preserved. Thus, nodes can be revisited if initially skipped, and this procedure is therefore guaranteed locate the target object. Once the target object is successfully seen, the search is finished.

\section{Generation of Plausible Scene Graphs}
\label{sec:scenegraphs}

To train and evaluate HMS, we generate 1500 scene graphs with plausible placements of storage locations within rooms, and dense placements of semantically related objects that may occlude the target on shelves. Existing simulation environments for evaluating search and navigation tasks~\cite{kolve2017ai2, xia2019interactive}, and annotated 3D Scene Graphs from real-world buildings~\cite{armeni20193d} both lack annotations for object storage locations and lack dense object arrangements with object occlusion. Thus, we propose a procedural generation process for creating synthetic but realistic scene graphs.

We propose that scene graph generation must produce plausible and varied hierarchies with dense placements of semantically related objects.
For example, in plausible and varied hierarchies, there is usually a fridge in the kitchen, but sometimes also a fridge in the garage or in rare cases a fridge in the living room.  These hierarchies need also have semantic clusters, in which objects are stored in containers together with semantically related objects, with some containers having two or more groups of related objects (e.g. a pantry having coffee and tea products on one side of a shelf, and protein products on the other side of the shelf).

To address the need of having plausible but varied hierarchies of containers, we focus on a hand-picked set of room and storage location types and manually set the probabilities of each storage location occurring in every type of room (e.g. a fridge is most likely to be in the kitchen, but has some probability of being in the living room).  Concretely, our procedure makes use of 7 room types (living room, closet, bedroom, garage, bathroom, closet) and 4 storage location types (fridge, cabinet, shelves, pantry). Since the same storage type can be in multiple rooms (e.g. shelves in both the bedroom and bathroom), this results in at most $4^7=16384$ possible combinations. Thus, to create many plausible but varied scene graphs we sample storage location placements from this distribution. We also associate each storage location with specific 3D models so as to have a known volume and number of shelves. 

To address the need of having plausible and dense placements of semantically related objects on the shelves of storage locations, we utilize a second procedural generation method. First, we manually curate 3D object models and categorize them into distinct groups based on their semantic identity. We also annotate each model with a probability per stable orientation for placements on a support surface, a text description (e.g. ``Handmade Brown Hat''), and a class label (e.g. ``Hat''). Each of the groups is annotated with the probabilities of being in a given storage location. Lastly, to place objects inside of containers we use the PyBullet~\cite{coumans2019} physics simulator to sample collision-free positions for all objects by sequentially dropping them and waiting for them to come to rest on a given shelf. To generate realistic-looking RGB images, we use the iGibson renderer~\cite{xia2019interactive}. 

We consider a single scene graph as the hierarchy of containers and the associated placements of objects within each storage location. Due to the variety of possible ways to assign storage locations to rooms and place objects on shelves, this procedure can generate a large number of plausible but varied scene graphs. We use this procedure as the basis for our experiments.

\section{Experiments}
\label{sec:experiments}

\begin{table}[!b]
\centering
\vspace{-15pt}
\caption{Accuracy Results}
\begin{tabular}{lcc}
\toprule
 & Container Accuracy & Object Accuracy \\
\midrule
Word Vector Similarity*            & 75                 & 56               \\
Nearest*        & -  &   64           \\ 
Largest*        & -  &   69              \\ 
Most Likely*    & 82    & -                \\ 
HMS, Object Context Vector & 90    & 69               \\ 
HMS, No Message Passing     & 76                 & 89               \\ 
HMS, No Object Label  &  91 &     86              \\ 
HMS, No Object Image  &  94 &     75              \\ 
HMS  & 92   & 89               \\ 
\bottomrule
\multicolumn{3}{L{0.95\linewidth}}{\scriptsize Average classification accuracies on the entire test set of scene graphs. Values are rounded to nearest integer. *has access to information not available to our model
}
\end{tabular}
\label{fig:table-accuracy}
\end{table}

\begin{table}[!t]
\centering
\caption{Search Efficiency Results}
\begin{tabular}{lcccccc}
\toprule
& Oracle* & (1)RND &  (4)NR* &  (5)LR* & (6)CV &  HMS  \\ 
\midrule
Oracle* & 4.3 & 7.4 &  5.8  & 7.0 & 7.3 &   5.4 \\
(1)RND & 72.5 & 76.8 &  81.0 &  75.0 & 79.2  &  73.5  \\
(2)WV* & 55.7   & 57.7   &  58.4 & 57.6   &   57.7     &   60.9  \\
(3)MS* & 15.3 & 17.2 & 16.0  &  17.5 &  18.5  &   15.6 \\
HMS & \textbf{13.6} & \textbf{14.5} &  \textbf{14.3} & \textbf{15.6} & \textbf{16.2}  &  \textbf{13.8} \\

\bottomrule
\multicolumn{7}{L{0.95\linewidth}}{\scriptsize Average test scene graph action counts (\# containers searched + \# objects moved). Rows correspond to container selection method, columns correspond to object selection method. *has access to information not available to our agent}
\end{tabular}
\label{fig:table-search}
\vspace{-15pt}
\end{table}

We evaluate HMS on 1500 synthetic scene graphs (Sec.~\ref{sec:scenegraphs}), and analyze its performance to address the following questions:
\begin{itemize}[wide, labelwidth=!, labelindent=0pt]
    \item How well does our approach perform object search?
    \item To what extent does integrating information across different levels of the scene graph guide our model?
    \item To what extent does our model leverage the visual input of objects to decide on occlusion probability? 
    \item  Does our model correctly prioritize semantically and geometrically appropriate objects when evaluating occlusion?
\end{itemize}

\paragraph*{Experimental Setup}
For our object set, we curate 135 3D object models from the Google Scanned Objects set~\cite{objects}, and 29 3D object models bought online, and group these models into 26 categories such a fruits, dairy products, hats, board games, etc. We use the same object models when generating both train and test scene graphs. Each non-empty shelf of a storage container has either 1 or 2 object categories and between 4 and 15 objects placed in plausible arrangements that often result in semantically related objects occluding each other. We generate 1000 unique scene graphs for training and 500 unique scene graphs for testing; train and test scene graphs have slightly different probabilities of container and object placement. When sampling target objects during training, we hold out 26 objects (1 from each category) to use as a search target at test time. We implement our model using the Pytorch framework~\cite{NEURIPS2019_9015}, and train for 50 epochs of 100 batches, sampling 10 scene graphs per batch. 

\paragraph*{Baselines and Ablations}
We compare to a set baselines, which all output a value in $[0,1]$ instead of $p$:
\begin{enumerate}[wide, labelwidth=!, labelindent=0pt]
    \item Random (RND): a uniformly random value in $[0,1]$ 
    \item  Word Vector Similarity (WV): the cosine similarity of the node word vector features and those of the target object. 
    \item  Most Likely (MS): the ground truth probability of a container having the class of the target object, as used in the procedural generation of scene graphs. 
    \item  Nearest (NR): how near the object is to the camera, normalized so that the farthest object on a shelf has a value of 0 and the nearest a value of 1.
    \item Largest (LR): the size of the object, normalized so that the smallest object on a shelf has a value of 0 and the largest has a value of 1.
    \item Object Context Vector (CV): similar to ~\cite{qiu2020target}, each object is associated with a 5-dimensional ``context vector'' which contains its bounding box information and word vector similarity to the target. We train a new model for which each object input representation is a context vector, and the model is otherwise the same as that of HMS.
\end{enumerate}

We also evaluate ablations of the model: no object labels, no object images, and no message passing.


\paragraph*{Metrics}
We focus on two metrics: classification accuracy and search efficiency. Classification accuracy is the measure of how often our model's classification network correctly predicts whether a container has the target object or an object occludes the target object. Different to this, search efficiency refers to the number of actions our approach requires to find the target object within a scene graph. Every decision with regards to a container or object is counted as one action. 
Because search involves reasoning about both containers and objects, we evaluate the search efficiency metric in terms of each possible combination of how containers are chosen and how objects are chosen. For search, the value of $T$ is set to $0.1$ initially for all nodes.

\begin{figure}[!h]

\vspace{-5pt}
\centering
    \includegraphics[width=0.95\linewidth]{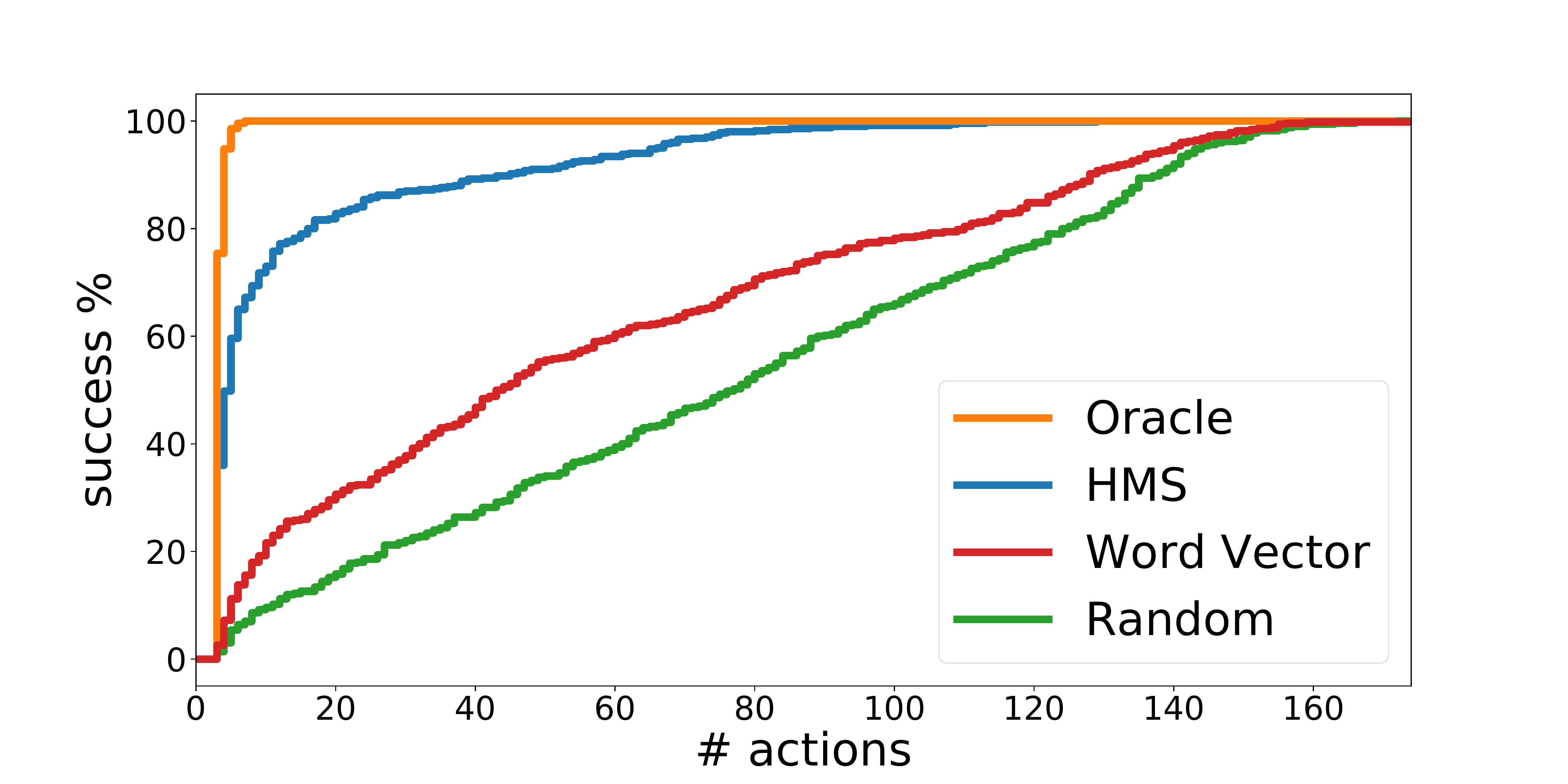}
   \caption{Percent of test set searches that succeeded in finding the target object in at most $x$ actions, for different search policies. HMS can find objects with significantly fewer actions compared to the baselines. The 'Word Vector' policy does object selection randomly.  } 
    
  \label{fig:dist}
  \vspace{-15pt}
\end{figure}

\paragraph*{Results}
Results in Table~\ref{fig:table-accuracy} suggests that the HMS approach outperforms baselines and ablations in accuracy by correctly predicting containers and occluders of the target object. Further, it relies on the image input for occlusion predictions.
Due to the higher accuracy in predictions, an agent using HMS finds the target object with significantly fewer actions when compared to the baselines (Table~\ref{fig:table-search}). While the average action count is higher than that of the oracle policy, this is primarily due to outlier scene graphs that require large action counts (Fig. \ref{fig:dist}). These scene graphs have rare container placements that are underrepresented in the training set. 

\begin{figure}[t!h]
    \centering
    \vspace{-5pt}
    \begin{subfigure}{80pt}
    \centering
    \includegraphics[width=\linewidth]{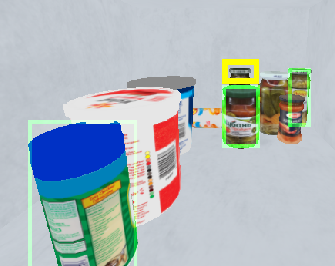}
    \end{subfigure}%
    \hfill
    \begin{subfigure}{80pt}
    \centering
    \includegraphics[width=\linewidth]{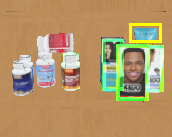}
    \end{subfigure}%
    \hfill
     \begin{subfigure}{80pt}
     \centering
     \includegraphics[width=\linewidth]{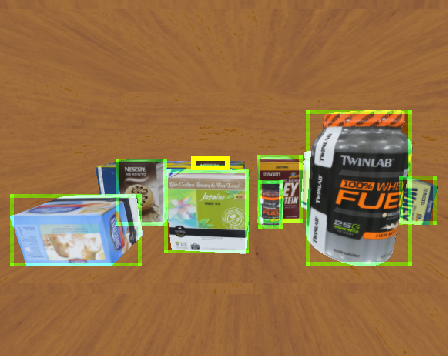}
     \end{subfigure}
    \caption{Qualitative examples of correct and flawed outputs by HMS. Yellow bounding boxes correspond to the visible portion of the target object, and greener bounding boxes correspond to higher model estimates that the object is occluding the target object. (left) Good predictions with target object \texttt{canned capers}: the model picks the more semantically relevant objects, despite the larger unrelated dairy products. (middle) Good predictions with target object \texttt{hair product}: the geometrically and semantically likely objects are preferred. (right) Flawed predictions with target object \texttt{nescafe coffee}: the products on the right are similarly scored as the more semantically relevant products on the left.
    }
  \label{fig:examples}
  \vspace{-15pt}
\end{figure}

HMS usually correctly combines semantic and geometric cues to select the correct occluding objects, but is occasionally incorrect.  Fig.~\ref{fig:examples} shows qualitative examples. However, flawed occlusion classification does not strongly effect search efficiency; the incorrect choice of container results in most of the unnecessary actions in searches.

\section{Conclusion} 
\label{sec:conclusion}

We presented \textit{Hierarchical Mechanical Search (HMS)}, an approach to search for a target object in organized indoors environments leveraging prior information about the layout of the environment in the form of 3D Scene Graphs, and reasoning jointly about semantics and geometry. Through a novel neural message passing architecture trained on 1000 3D scene graphs, and a novel search procedure, HMS recursively reasons across layers of the graph to help an interactive agent to find a target object specified with natural language descriptions. Evaluation on 500 3D scene graphs with dense placements of semantically related objects shows that HMS is significantly superior to several baselines at finding objects efficiently. In the future, we will work to make a non-synthetic dataset for this problem, evaluate HMS in real indoor environments, and extend HMS with low level control implementations for the desired navigation and manipulation actions on containers and objects~\cite{martin2017cross,danielczuk2019mechanical,kurenkov2020visuomotor}.


\renewcommand*{\bibfont}{\footnotesize}
\printbibliography 

\end{document}